\documentclass[11pt]{article}
\usepackage{nodalida2021}
\usepackage{times}
\usepackage{url}
\usepackage{latexsym}
\usepackage{hhline}
\usepackage{graphicx}
\usepackage{epstopdf} 

\aclfinalcopy % Uncomment this line for the final submission
\def\aclpaperid{***} %  Enter the acl Paper ID here

\usepackage[T1]{fontenc}
\usepackage[utf8]{inputenc}

\usepackage{microtype}

\usepackage{CJKutf8}

\usepackage[russian,english]{babel}

\DeclareRobustCommand{\cyrins}[1]{%
  \begingroup\fontfamily{cmr}%
  \foreignlanguage{russian}{#1}%
  \endgroup
}

\title{Grammatical Error Generation Based on Translated Fragments}

\author{Eetu Sjöblom \and Mathias Creutz \and Teemu Vahtola \\
  Department of Digital Humanities, Faculty of Arts, University of Helsinki, Finland \\
  \texttt{\{eetu.sjoblom,mathias.creutz,teemu.vahtola\}@helsinki.fi}}

\begin{document}
\maketitle
\begin{abstract}
We perform neural machine translation of sentence fragments in order to create large amounts of training data for English grammatical error correction. Our method aims at simulating mistakes made by second language learners, and produces a wider range of non-native style language in comparison to state-of-the-art synthetic data creation methods. In addition to purely grammatical errors, our approach generates other types of errors, such as lexical errors. We perform grammatical error correction experiments using neural sequence-to-sequence models, and carry out quantitative and qualitative evaluation. A model trained on data created using our proposed method is shown to outperform a baseline model on test data with a high proportion of errors.

% ... Our method aims at simulating mistakes made by second language learners, and produces a wider range of non-native style language in comparison to a state-of-the-art baseline model. We carry out quantitative and qualitative evaluation. Our method is shown to outperform the baseline on data with a high proportion of errors.
\end{abstract}

\section{Introduction}

Grammatical error correction (GEC) is the task of detecting and correcting grammatical errors in texts, typically written by second language learners. Current state-of-the-art GEC approaches are based on neural machine translation (NMT) \cite{grundkiewicz-etal-2019-neural}. As in other natural language processing tasks, neural approaches to GEC rely on large quantities of task-specific data, that is, sentence pairs consisting of erroneous source text coupled with corrected target text. However, in-domain GEC data is scarce, and a number of solutions to the data sparsity problem have been proposed recently, often by introducing artificially created GEC data into the training process.

Some error generation approaches also depend on error-annotated authentic learner data. For example, \citet{felice-yuan-2014-generating} introduce errors probabilistically with error probabilities that are estimated using a learner corpus. \citet{rozovskaya-etal-2014-correcting} train error detection and classification models on annotated data, focusing on verb errors. Other methods dispense with the need for annotated data, such as approaches based on inverted spell-checkers and heuristic error generation \cite{grundkiewicz-etal-2019-neural, grundkiewicz-junczys-dowmunt-2019-minimally}.  

To alleviate the data sparsity problem, in this work we propose to use NMT to produce artificial training data, simulating real errors made by language learners. For instance, to produce English text with errors, we use NMT models to translate sentence fragments from other languages to English, and then combine the translated fragments to form our erroneous source data. Similar machine translation approaches to GEC data creation have been proposed before. For example, \citet{rei-etal-2017-artificial} use a statistical machine translation model trained on reversed learner data, using the corrected sentences as source data and erroneous sentences as targets. \citet{kasewa-etal-2018-wronging} extend this approach and use an NMT model to produce errors. \citet{htut-tetreault-2019-unbearable} perform extensive experiments on several neural models, likewise trained on learner data to generate errors.

Our contribution is to split the foreign-language source sentences into shorter fragments in order to limit the context that is available to the machine translation system. The rationale for doing this is to produce text that contains artefacts from the foreign language. Since the NMT system needs to translate shorter fragments without the proper context, we expect it to produce more literal translations and to be less able to produce correct agreement between different parts of speech. Additionally, polysemy may prompt the system to suggest translations of a synonym in the foreign language, which is not a synonym in English. The creation of synthetic training data involves further steps, which are described in Section~\ref{sec:training_data}. Model training is explained in Section~\ref{sec:model}. In Section~\ref{sec:evaluation} we evaluate our approach quantitatively against a strong baseline \cite{grundkiewicz-etal-2019-neural} and make some qualitative assessments.

\section{Training data}
\label{sec:training_data}

The creation of our training data involves the following steps:

\begin{enumerate}
    
    \item English sentences aligned with sentences of other languages are used as data.\footnote{These languages, which have been chosen to represent both European and Asian languages from different language families are the following: Danish, Dutch, Finnish, French, German, Italian, Japanese, Korean, Latvian, Portuguese, Russian, Spanish, and Swedish.} Our parallel text data are retrieved from the OpenSubtitles \cite{lison2016lrec} and Europarl \cite{koehn2005mtsummit,TIEDEMANN12.463} collections.\footnote{Available for download at: \url{https://opus.nlpl.eu/}} 
    
    \item The non-English sentences are split randomly into chunks of an average length of three word tokens.
    
    \item Each sentence chunk in isolation is translated into English using OPUS-MT machine translation models from HuggingFace \cite{TiedemannThottingal:EAMT2020}. N-best lists containing up to ten alternate translations for each chunk are produced.
    
    \item Full English sentences are created by concatenating chunks from the n-best lists. Ten different alternate full sentence translations are generated for each source sentence by combining chunks at random, proportionally to the translation scores of the chunks. Our aim is to obtain English translations that contain errors influenced by the source language. The original English sentence from the parallel corpus serves as a correct reference translation. Examples are shown in Table~\ref{tab:broken_transl}. 
    
    \item In theory, for each sentence in our data we now have ten artificially created, erroneous English sentences. However, many of the synthetic sentences do not resemble authentic human-produced erroneous sentences. We therefore discard a significant portion (60\,\%) of the synthesized sentences by sampling for an error distribution that is closer to the error distribution of authentic data, represented by our development sets. This leaves us with just 23\,\% of the words of our original set, reflecting the fact that longer sentences are more likely to be discarded.
    
    \item The above mentioned sampling of sentences requires us to be able to compare error distributions between authentic and synthetic data. First we POS tag the sentence pairs and align them automatically using minimum string edit distance coupled with some heuristics taking into account part of speech and inflection. The alignment algorithm is similar but not identical to ERRANT \cite{Bryant2017acl, Felice2016coling}. This procedure is illustrated in Table~\ref{tab:corr_probs}. From the alignments we extract trigrams consisting of a correction operation in the context of one preceding and following token, such as \textit{PRP ins(MD) VBP} (``insert a modal verb between a pronoun and a verb in non-third person singular form'') or \textit{ins(DT) ins(JJ) NN} (``insert an adjective between an inserted determiner and a noun in singular''). These automatically extracted trigrams constitute our correction types. Their frequency distributions are not the same across the authentic and synthetic data. We filter the synthetic data by keeping sentence pairs that contain combinations of correction types that are highly likely to occur in authentic data and discard sentence pairs with low-probability correction types.

\end{enumerate}

\begin{table*}[t]
\begin{small}
\begin{tabular}{l|p{0.9\textwidth}}

de-src &  W\"ahrend / du / bewusstlos im / Krankenhaus / lagst, sind / die \"Arzte mit / diesen / Testergebnissen / zu / Daniel gekommen.\\
en-tgt & \textbf{During} / \textbf{you} / \textbf{unconscious in} / \textbf{Hospital} / \textbf{the} / \textbf{doctors with} / \textbf{the} / \textbf{Test results} / \textbf{to} / \textbf{Daniel came.} \\
en-ref &  While you were unconscious in the hospital, the doctors came to Daniel with these test results.\\
\hline
ru-src & \cyrins{И никогда не} / \cyrins{переставал} / \cyrins{думать о} / \cyrins{тебе.}\\
en-tgt & \textbf{And never} / \textbf{I stopped} / \textbf{to think about} / \textbf{You.}\\
en-ref & I never stopped thinking about you.\\
\hline
fr-src & Il est / vrai que toutes les / histoires ne peuvent avoir une fin heureuse, mais pour Jules / Daly, la r\^eveuse de Buffalo, / l'histoire ne / fait que commencer.\\
en-tgt & \textbf{It is} / \textbf{true that all} / \textbf{stories can't have a happy ending, but for Jules} / \textbf{Daly, Buffalo's dreamer,} / \textbf{history} / \textbf{Just start.}\\
en-ref & It is true not all tales have happy endings, but then for Jules Daly, the dreamer from Buffalo, the story is just beginning.\\
\hline
ja-src & \begin{CJK}{UTF8}{ipxm}もし君が生き残れたら\end{CJK} / \begin{CJK}{UTF8}{ipxm}一生懸命に働いたからだ\end{CJK} \\
en-tgt & \textbf{And if you survive,} / \textbf{Because you worked hard.}\\
en-ref & If you live, you have worked very hard indeed.\\
\hline
fi-src & Koulu / on - / l\"ahett\"anyt / minut useammalle / terapeutille kuin / sinulla / on ollut huonoja / treffej\"a.\\
en-tgt & \textbf{School} / \textbf{is} / \textbf{sent} / \textbf{me to more} / \textbf{for a therapist} / \textbf{you} / \textbf{has been bad} / \textbf{Date.}\\
en-ref & This school has sent me to more therapists than you've had bad dates.\\
\end{tabular}
\end{small}
\caption{Sentences in other languages (*-src) are split into chunks (e.g., / \textit{bewusstlos im} /), and each chunk in isolation  is translated automatically into English. By concatenating the chunks we obtain English sentences containing errors (en-tgt), for which correction hypotheses exist in the form of the English reference translations (en-ref). }
\label{tab:broken_transl}
\end{table*}

\begin{table*}[t]
%\begin{small}
\begin{tabular}{l l}
\hline
Learner sentence: & We had enjoy time . \\
Correction: & We had a great time . \\
Alignment: & PRP VBD del(VB) ins(DT) ins(JJ) NN . \\
\hline
Synthetic sentence: & You be the the old donkey of the forestry \\
Correct reference: & You 'll be the oldest donkey in the forest . \\
Alignment: & PRP ins(MD) VBP del(DT) DT inf(JJS) NN del(IN) ins(IN) DT typ(NN) ins(.)\\
\hline
\end{tabular}
%\end{small}
\caption{Pairs of sentences with alignments. The upper example is an authentic sentence produced by a language learner accompanied by a correction (target hypothesis) proposed by a teacher. The alignment is a sequence describing how to modify the learner sentence into the corrected one. It reads as follows: \emph{PRP}: keep pronoun (``we''), \emph{VBD}: keep verb in past tense (``had''), \emph{del(VB)}: delete verb in infinitive (``enjoy''), \emph{ins(DT)}: insert determiner (``a''), \emph{ins(JJ)}: insert adjective (``great''), \emph{NN}: keep noun in singular (``time''), \emph{.}: keep punctuation. The lower example is analogous, but the alignment is between a synthetically produced sentence and the correct English reference. This alignment sequence contains a few more correction types: \emph{inf(JJS)}: change inflection of adjective into superlative (``oldest''), \emph{typ(NN)}: fix typo in noun in singular (the word ``forestry'' is here classified as a spelling error by the algorithm).}
\label{tab:corr_probs}
\end{table*}

\subsection{Final training sets}

We carry out experiments using systems trained on four different training sets. We create one data set using our method that matches the word count of the Baseline comparison. In addition, we create two smaller data sets using both the Baseline method and ours on the same correct target sentences in order to control the effects of data domain.

\begin{itemize}
    % \item \textbf{Baseline:} We compare our own training scheme to a system trained on the training set created by \citet{grundkiewicz-etal-2019-neural}, consisting of 100 million sentence pairs. They propose an unsupervised data generation method based on confusion sets from spellcheckers. For each source sentence in the training data, they replace a random number of tokens with a substitute from the vocabulary item's confusion set. In addition, they probabilistically delete and insert random tokens, as well as swap adjacent tokens in the sentence. Using similar operations at the character level they introduce additional noise to make their models more robust to spelling errors. We expect models trained on this data to perform well on learner data that contains errors mostly at the level of single tokens. Although swaps, insertions, and deletions introduce some syntactic and word order mistakes, the method does not excel at producing more complex syntactic errors, errors that require extensive reordering of the sentence, or errors that result from L1 influence.
    \item \textbf{Baseline:} We compare our own training scheme to a system trained on the training set created by \citet{grundkiewicz-etal-2019-neural}. They propose an unsupervised data generation method based on confusion sets from spellcheckers. For each source sentence in the news crawl data used for training, they replace a random number of tokens with a substitute from the vocabulary item's confusion set. In addition, they probabilistically delete and insert random tokens, as well as swap adjacent tokens in the sentence. They also introduce additional noise at the character level using similar operations. Although these operations introduce some syntactic and word order mistakes, the method does not excel at producing more complex syntactic errors, errors that require extensive reordering of the sentence, or errors that result from L1 influence.
    
    %BRIEFLY CHARACTERIZE HOW THEY SIMULATE ERRORS AND POSSIBLE IMPLICATIONS. THIS PAPER NEEDS TO EXPLAIN THE MAIN DIFFERENCE BETWEEN THE TWO MODEL TYPES. OTHERWISE OUR READERS WILL NOT UNDERSTAND WHAT WE ARE DOING.
    
    %\item \textbf{Sentence-Match:} We produce a training set using our method, which is sampled to contain the same number of sentence pairs as the Baseline.
    
    %\item \textbf{Word-Match:} Since sentences produced by our method are shorter on average than in the Baseline model, we sample another, larger training set where the number of word tokens match with that of the Baseline (4.6 billion words).
    
    \item \textbf{Chunks:} We produce a training set using our method, which is sampled to contain the same number of words as the Baseline (4.6 billion words).
    
    \item \textbf{Chunks-small:} We produce a training set using our method such that the data set contains only unique target sentences. This smaller set contains approximately 650 million word tokens and allows for faster model training.
    
    \item \textbf{Baseline-small:} We use the Baseline data creation method on the same target sentences as in the Chunks-small set.
    
    %\item \textbf{Mixed:} Finally we create a training set consisting of the Sentence-Match set (1.5 billion words) combined with a subset of the Baseline set (3 billion words). This set contains as many word tokens as the Baseline and the Word-Match sets.
\end{itemize}

\section{Model training}
\label{sec:model}
% We run our experiments with large Transformer models, which have achieved state-of-the-art performance on GEC \cite{grundkiewicz-etal-2019-neural}. 
We build on the system described in \citet{grundkiewicz-etal-2019-neural}. We choose not to make changes to the model or training parameters in order to isolate the effects of our data creation method and ensure a fair comparison. The same training setup is used for all models, with modifications only in the training sets. Specifically, we use their ``Transformer Big'' architecture, with 6 self-attention layers, 16 attention heads, embeddings vectors of size 1024, and feed-forward hidden size of 4096 with ReLU activation functions. We also tie the encoder, decoder, and output embeddings.

We also adopt the training setup of \citet{grundkiewicz-etal-2019-neural}, and train our models with the Marian toolkit \cite{junczys-dowmunt-etal-2018-marian}. The models are first pretrained on the synthetic data for a maximum of 5 epochs. After pretraining, we finetune the best model checkpoint using the following corpora: FCE \cite{yannakoudakis-etal-2011-new}, NUCLE \cite{dahlmeier-etal-2013-building}, W\&I-LOCNESS \cite{bryant-etal-2019-bea,granger1998computer}, and Lang-8 \cite{mizumoto-etal-2012-effect}. We use the W\&I-LOCNESS development set for validation during training.

We use early stopping with a patience of 10 with ERRANT $F_{0.5}$ score on the W\&I+locness development set used as the early stopping criterion. The checkpoint with the highest $F_{0.5}$ score is chosen for further finetuning. We choose the ADAM optimizer, a learning rate of 0.0002 and a linear warmup for 8k updates. We use Marian's option to dynamically fit mini-batches to GPU memory, and train our models using 4 Nvidia Volta V100 GPUs (32GB RAM). In addition, we use strong regularization, which has been found useful in GEC systems, with dropout probabilities of 0.3 between layers, 0.1 for self-attention and feed-forward layers, 0.3 for entire source token embeddings, and 0.1 for target embeddings.

\section{Evaluation}
\label{sec:evaluation}

% ORIGINAL
% \begin{table}
%     \centering
%     \begin{tabular}{|l|c|c|}
%         \hline
%          & W\&I-LOCNESS & YKI \\
%          \hline
%          Baseline & \textbf{66.44} & 52.63 \\
%          Sentence-Match & 65.12 & 49.98 \\
%          Word-Match & 65.44 & \textbf{53.41} \\
%          Mixed & 65.13 & 52.41 \\
%          \hline
%     \end{tabular}
%     \caption{$F_{0.5}$ scores for the four models on the two test sets. $F_{0.5}$ is a weighted harmonic mean of precision and recall, where precision is accentuated.}
%     \label{tab:testscores}
% \end{table}

\begin{table}
    \centering
    \begin{tabular}{|l|c|c|}
        \hline
         & W\&I-LOCNESS & YKI \\
         \hline
         Baseline & \textbf{66.44} & 52.63 \\
         Chunks & 65.44 & \textbf{53.41} \\ \hhline{|=|=|=|}
         Baseline-small & \textit{60.09} & 46.66 \\
         Chunks-small & 59.89 & \textit{49.73} \\
         \hline
    \end{tabular}
    \caption{$F_{0.5}$ scores for the four models on the two test sets. $F_{0.5}$ is a weighted harmonic mean of precision and recall, where precision is accentuated.}
    \label{tab:testscores}
\end{table}

% Where do we explain how yki was created and why?
We report results on two different data sets. We use the W\&I-LOCNESS set, which was used as official test data in the BEA19 GEC shared task \cite{bryant-etal-2019-bea}, as well as a subset of the English portion of learner texts derived from the Finnish National Certificates of Language Proficiency exams (YKI).\footnote{Available for research purposes from the Centre for Applied Language Studies at the University of Jyv\"askyla, Finland: \url{http://yki-korpus.jyu.fi/}} We do not use the YKI data as a blind test set, but instead use it to qualitatively analyze differences in model predictions. Still, no part of the YKI data was used during training or development of the models.

We compare our results with those reported by \citet{grundkiewicz-etal-2019-neural}, whose system achieved first place in the BEA19 GEC shared task. However due to limited resources we do not train an ensemble of models, but instead take a single left-to-right model from \citet{grundkiewicz-etal-2019-neural} as baseline. Their best system uses an ensemble approach with right-to-left and language model reranking and achieves a higher $F_{0.5}$ score of 69.47 on the W\&I-LOCNESS test set.

The upper part of Table~\ref{tab:testscores} compares our Chunks model with the baseline by \citet{grundkiewicz-etal-2019-neural}. The Baseline model performs best on W\&I-LOCNESS with a one absolute point difference compared to our model. However, our model outpeforms the Baseline on YKI. These results suggest that our data creation method might be suitable when correcting noisier source sentences, as YKI generally contains more challenging language with more errors than W\&I-LOCNESS. 

%The Baseline and Word-Match models are trained on different data domains. To control the domain and better understand the differences due to the data augmentation methods, we apply both methods on the same target sentences to create two smaller training sets in the same domain. We use all unique target sentences from our data set based on OpenSubtitles and Europarl to create these data. 

The lower part of Table~\ref{tab:testscores} demonstrates that the trends are the same for the smaller models, in which we match the data domain in training. That is, the Baseline is no longer trained on news data but on OpenSubtitles and Europarl. The results are lower overall due to the smaller data size. The Baseline outperforms our model on W\&I-LOCNESS also in this setting, although by a smaller margin. However, the performance gap on YKI increases by approximately two absolute points in favor of our model, offering additional support that our method can improve performance on noisy data. To better understand differences between the models, we examine their predictions on the same source sentences, as described in the next section.

% The Sentence-Match model predictably performs worst, as it is pretrained on the least amount of data. Curiously, the Mixed model performs worse that the Word-Match model on YKI, that is, replacing some of the data used in training the Baseline model with our data decreases performance, but using a token-count-matched sample of only our data performs best. 

\subsection{Qualitative assessment}

% Precision and recall:
% Non-finetuned models:
% Small-nmt has significantly higher recall on both wiloc and yki sets.
% Small-nmt has lower precision on wiloc, and a little lower on yki.
% Finetuned models:
% Small-nmt has higher recall on both wiloc and yki.
% Small-nmt has only a little lower precision on wiloc but higher precision on yki.

% Qualitative assessment done on SMALL-FINETUNED models.
% Results are very similar
% Some differences between the reported Baseline and Word-Match and small models:
% Small-nmt does not correct higing to hiking, but small-gk does. --> contrary to what was reported on the bigger models, where Word-Match corrected but Baseline did not. ??
% wery and broblems not corrected by either model.
% Small-nmt corrects for example Agust --> August and diffiicul --> difficult, small-gk leaves them unchanged.
% Both succeed in correcting prize --> price

% We have taken a closer look at the corrections made by the models on the 320 sentences in the YKI data set. The results are surprisingly similar although the models (Chunks and Baseline) have been trained on different text corpora, into which errors have been introduced using different methods. Word-Match suggests slightly more corrections on average than the Baseline, yielding a somewhat higher recall and lower precision.
We have taken a closer look at the corrections made by the Baseline and Chunks models on the 320 sentences in the YKI data set. The results are surprisingly similar although the models have been trained on different text corpora, into which errors have been introduced using different methods. The Chunks model suggests slightly more corrections on average than the Baseline, yielding a somewhat higher recall and lower precision.
%The models trained on our data suggest slightly more %corrections on average, yielding a somewhat higher recall %and lower precision.

It is interesting to see that the Chunks model performs quite well on misspelled words (\textit{broblems}, \textit{i'ts}, \textit{beatyfull}) although it has not been explicitly trained to correct spelling mistakes, in contrast to the Baseline method. In the training data of the Baseline, spelling errors have been introduced by random sampling, whereas the models based on machine translated data generally do not contain any spelling mistakes, as machine translation does not generate them. Yet, it appears that the Chunks model corrects spelling errors at least as well, if not better, than the Baseline. The latter model leaves word forms, such as \textit{wery}, \textit{nicier} and \textit{higing} (for \textit{hiking}) unchanged.
% Tarvitaanko tällaista keskustelua lainkaan, pidentää osiota?: There are some differences between the larger models and the small ones. Our small model corrects words such as \textit{Agust} and \textit{diffiicul} while the small gk model leaves them unchanged. However, in some occasions, the small gk model corrects words that our model does not correct (\textit{higing}). Introducing more data to the models, interestingly, changes this into a contrary: \textit{higing} is corrected by the Word-Match but not by the Baseline.

When it comes to choosing the correct spelling in context, the Chunks model distinguishes between the different usages of \textit{prize} and \textit{price} (``\textit{The prize for you is betveen 1500-1700 euros.}''), and it has in fact been trained on almost 4000 sentence pairs in which \textit{prize} is corrected to \textit{price} in context. The Baseline does not make this correction.
% Surprisingly(?), both models trained on the small data make the correction.

% Neither model manages 
None of the models manage to correct the sentence ``\textit{I have old but wery fine cun selling.}''. Firstly, the models fail to change \textit{cun} into \textit{gun}. Secondly, one could have expected the Chunks model to see the connection between \textit{selling} and \textit{for sale}, since there are 800 training examples containing that substitution, but for some reason this particular test sentence does not trigger the desired change.

Many of the sentences in the YKI corpus are indeed hard to interpret without broader world knowledge; The Chunks model corrects the sentence ``\textit{If it isn't help then you will ask for help to polishman}'' into ``\textit{If it doesn't help then you will ask the Polishman for help.}''. However, the correct person to ask for help here would be the police man. In another sentence, ``\textit{I begin hobbies about 12 yers.}'', the model would somehow need to understand that the person picked up hobbies at the age of twelve rather than twelve years ago.

Additionally, we have examined the W\&I-LOCNESS development set, although it has been used as a stopping criterion in the training, which may bias the results slightly. The $F_{0.5}$ scores on the dev set are the same for both the Baseline and the Chunks model (52.6\,\%). This is considerably lower than for the final test set (65.4 - 66.4\,\%), suggesting that the test set is less challenging than the dev set. Compared to the YKI data, even the W\&I-LOCNESS dev set seems cleaner and appears to contain fewer mistakes. It is hard to see significant differences in performance between the models. For 65\,\% of the sentences, the Baseline and the Chunks model produce exactly the same corrections. The corresponding figure for the YKI set is 57\,\%.

\section{Discussion and conclusion}

We have shown that our model rivals a competitive baseline, a left-to-right model by \citet{grundkiewicz-etal-2019-neural}, which was one component of an ensemble model that performed best in the BEA19 GEC shared task. We did not yet train our own ensemble model, but we expect to see similar improvements in performance in future experiments.

Our results show that two models can perform on par, although they have been pretrained on different training corpora and using different error simulation techniques. In addition, the Chunks model outperforms the Baseline in noisy conditions. In the future, we would like to analyze further techniques for modeling challenging types of errors, which originate from structures that differ between the target language and the native languages of the language learners.

% Our experiments show that two models (Baseline vs. Word-Match) can obtain very similar results, although they have been pretrained on different training corpora and using different error simulation techniques. In the future, we would like to analyze further techniques for modeling challenging types of errors, which originate from  structures that differ between the target language and the native languages of the language learners. 

\section{Acknowledgments}
\vspace{1ex}
\noindent
\begin{minipage}{0.1\linewidth}
    \vspace{-10pt}
    \raisebox{-0.2\height}{\includegraphics[trim =32mm 55mm 30mm 5mm, clip, scale=0.2]{erc.ai}} \\[0.25cm]
    \raisebox{-0.25\height}{\includegraphics[trim =0mm 5mm 5mm 2mm,clip,scale=0.078]{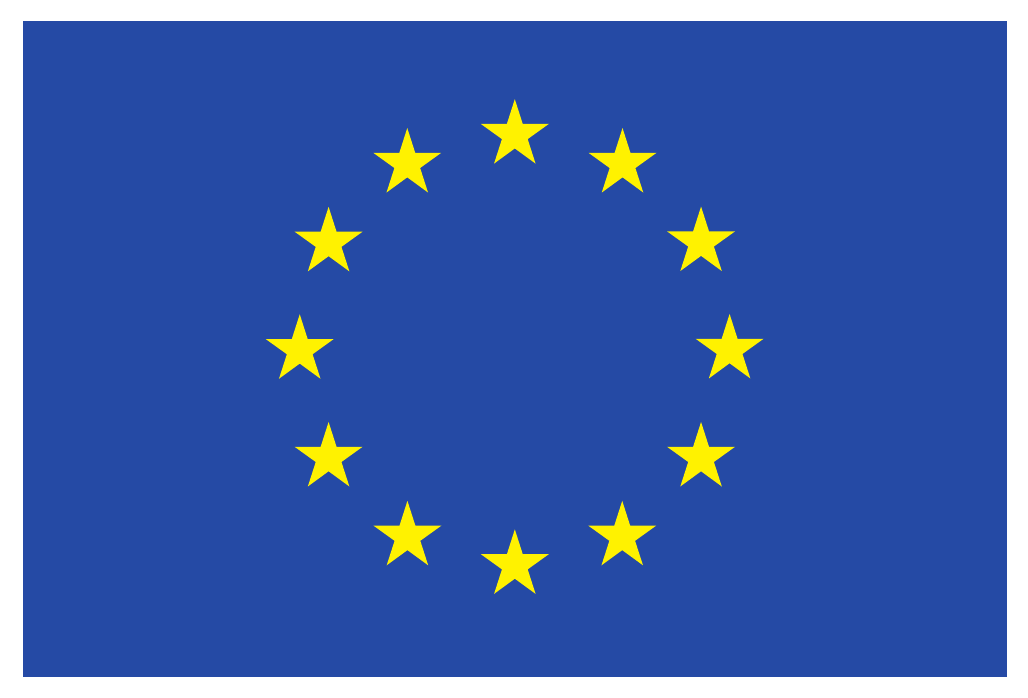}}
\end{minipage}
\hspace{0.01\linewidth}
\begin{minipage}{0.85\linewidth}
This study has been supported by the FoTran project, funded by the European Research Council (ERC) under the European Union's Horizon 2020 research and innovation programme (grant agree-\linebreak
  \vspace{-.7em}
\end{minipage}\linebreak ment \textnumero{}~771113).
We wish to acknowledge CSC – IT Center for Science, Finland, for generous computational resources.

\bibliographystyle{acl_natbib}
\bibliography{anthology,custom}

%\appendix

% \section{Example Appendix}
%\label{sec:appendix}

%This is an appendix.

\end{document}